\documentclass{article}

\PassOptionsToPackage{numbers, compress}{natbib}

\usepackage[final]{icbinb2021}

\usepackage[utf8]{inputenc} 
\usepackage[T1]{fontenc}    
\usepackage{hyperref}       
\usepackage{url}            
\usepackage{booktabs}       
\usepackage{amsfonts}       
\usepackage{nicefrac}       
\usepackage{microtype}      
\usepackage[table]{xcolor}  

\usepackage{amsmath}        
\usepackage{acro}           
\usepackage{multirow}
\usepackage{tabularx}
\usepackage{xspace}         
\usepackage{graphicx}
\usepackage{placeins}
\usepackage{pgfplots}

\newcommand{\figref}[1]{Fig.~\ref{#1}}

\newcommand{\eqnref}[1]{Eq.~\eqref{#1}}
\newcommand{\tabref}[1]{Table~\ref{#1}}

\makeatletter
\DeclareRobustCommand\onedot{\futurelet\@let@token\@onedot}
\def\@onedot{\ifx\@let@token.\else.\null\fi\xspace}
 
\def\ie{i.e\onedot}

\def\etal{et~al\onedot}

\makeatother

\DeclareAcronym{cnn}{short=CNN, long=convolutional neural network}
\DeclareAcronym{slc}{short=SLC, long=semi local convolution}

\title{Semi-Local Convolutions for LiDAR Scan Processing}

\author{
Larissa T. Triess$^{1,2}$
\And
David Peter$^{1,3}$\thanks{This work was conducted while David Peter was with Mercedes-Benz AG.}
\And
J. Marius Z\"ollner$^{2,4}$
\And
\\
$^{1}$Mercedes-Benz AG, Research and Development, Stuttgart, Germany\\
$^{2}$Karlsruhe Institute of Technology, Karlsruhe, Germany\\
$^{3}$Robert Bosch GmbH, Stuttgart, Germany\\
$^{4}$Research Center for Information Technology, Karlsruhe, Germany\\
\\
\texttt{larissa.triess@daimler.com}
}

\begin{document}

\maketitle


\begin{abstract}
    A number of applications, such as mobile robots or automated vehicles, use LiDAR sensors to obtain detailed information about their three-dimensional surroundings.
    Many methods use image-like projections to efficiently process these LiDAR measurements and use deep convolutional neural networks to predict semantic classes for each point in the scan.
    The spatial stationary assumption enables the usage of convolutions.
    However, LiDAR scans exhibit large differences in appearance over the vertical axis.
    Therefore, we propose \ac{slc}, a convolution layer with reduced amount of weight-sharing along the vertical dimension.
    We are first to investigate the usage of such a layer independent of any other model changes.
    Our experiments did not show any improvement over traditional convolution layers in terms of segmentation IoU or accuracy.
\end{abstract}


\section{Introduction}
\label{sec:introduction}

For a wide variety of applications it is important to understand the semantic meaning of the three dimensional world.
LiDAR sensors provide a precise sampling of the environment and are often used for object detection and semantic segmentation.
Many state-of-the-art semantic segmentation approaches make use of traditional two-dimensional \acp{cnn} by projecting the point clouds into an image-like structure~\cite{Milioto2019IROS,Wu2019ICRA}.

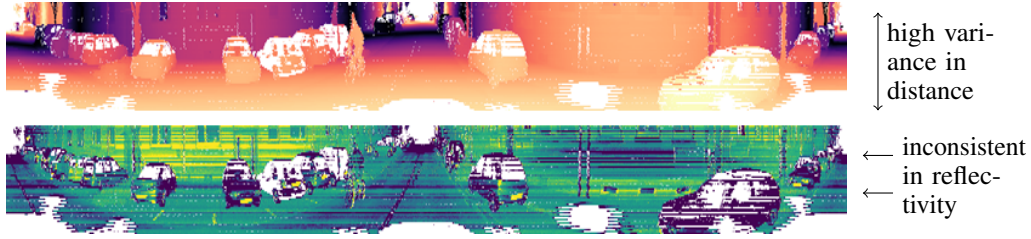
\begin{figure}
    \centering
    \begin{tikzpicture}
    \node at (0,0) {\includegraphics[width=0.8\linewidth]{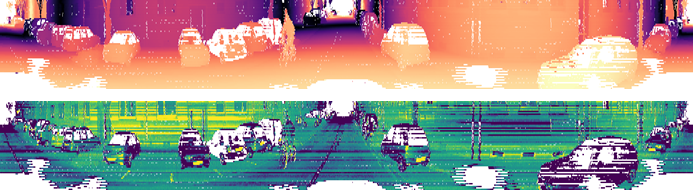}};
    \draw[<->] (6,0.1) -- (6,1.4) node[anchor=west, pos=0.5, text width=1.4cm] {high variance in distance};
    \draw[<-] (5.8, -0.5) -- (6.2, -0.5);
    \draw[<-] (5.8, -1.0) -- (6.2, -1.0);
    \node[anchor=west, text width=1.4cm] at (6.2,-0.8) {inconsistent in reflectivity};
    \end{tikzpicture}
    \caption{
        \textbf{LiDAR Scan}:
        A projected LiDAR scan has a high variance in distance over the vertical axis of the scan (top) and some sensors have calibration issues of the reflectivity module that varies for each layer (bottom).
    }
    \label{fig:scan}
\end{figure}

In this paper we provide deeper insights into one of the methods developed in our previous work~\cite{TriessIV2020}.
We investigate whether the spatial stationary assumption of convolutions is still applicable to inputs with varying statistical properties over parts of the data, such as projected LiDAR scans.
These data structures exhibit similar features as aligned images for which locally connected layers have been introduced~\cite{Taigman2014CVPR}.
\figref{fig:scan}~shows an example of such a projected LiDAR scan, which has a high variance in depth over the vertical axis, but not the horizontal axis.
Further, the reflecivity measurements show inconsistencies over the vertical axis.
To address this effect, we introduce \ac{slc}, a layer with reduced weight-sharing along the vertical spatial dimension.


\section{Related Work}
\label{sec:related}

Convolution layers apply a filter bank on their input.
The filter weights are shared over all spatial dimensions, meaning that for every location in the feature map the same set of filters are learned.
The re-usability of weights causes a significant reduction in the number of parameters compared to fully connected layers.
This allows deep convolutional neural networks to be trained successfully, in turn leading to a substantial performance boost in many computer vision applications.
The underlying premise of convolutional methods is that of translational symmetry, \ie that features that have been learned in one region of the image are useful in other regions as well.

For applications such as face recognition which deal with aligned data, locally connected layers have proved to be advantageous~\cite{Gregor2010,Huang2012CVPR,Taigman2014CVPR}.
These layers also apply a filter bank.
Contrary to convolutional layers, weights are not shared among the different locations in the feature map, allowing different sets of filters to be learned for every location in the input.

The spatial stationary assumption of convolutions does not hold for aligned images due to different local statistics in distant regions of the image.
In a projected LiDAR scan, the argument holds true for sensors that are mounted horizontally.
Each horizontal layer is fixed at a certain vertical angle.
As the environment of the sensor is not invariant against rotations around this axis, this leads to different distance statistics in each vertical layer.

There exists some other works that exploit variability in depth along the vertical axis~\cite{Bewley2020CORL,Brazil2019ICCV,Ding2020CVPR,Fan2021ICCV,Sirohi2021}.
Though their work is conceptually similar to ours, they report improvements in contrast to this work.
However, these works do not include an ablation study on this particular architectural modification independent of the other modifications proposed in their work.
This makes the evaluation not transparent about the actual influence of the adapted convolutions.
Therefore, this work presents the results of an independent ablation and shows that explicit modelling of varying statistical properties in \acp{cnn} is not necessary for LiDAR-base semantic segmentation.

\section{Method}
\label{sec:method}

We consider an input feature map~$x$ of shape~$\left[H_x,W_x,C_x\right]$, representing a cylindrical projection with height~$H_x$, width~$W_x$, and channels~$C_x$.
It is passed through a convolution layer which outputs another feature map~$y$ of shape~$\left[H_y,W_y,C_y\right]$.

A conventional 2D convolution layer computes the output~$y$ with a cross-correlation of the input~$x$ and a filter kernel~$k$ such that
\begin{equation}
\label{eq:conv}
    y_{h, w, c_y} =
    \sum_{c_x=0}^{C_x-1} \sum_{i=0}^{I-1} \sum_{j=0}^{J-1}
    k_{i, j, c_x, c_y} \cdot
    x_{h+i-I, w+j-J, c_x}
\end{equation}
where $k$ has the shape~$\left[2I+1,2J+1,C_x,C_y\right]$.

In a \ac{slc} layer, the filter kernel has the shape~$\left[2I+1,2J+1,C_x,C_y,\alpha\right]$ where $\alpha$ is the number of components with~$\{\alpha\in\mathbb{N}: 1 \leq \alpha \leq H_x\}$.
Each component is responsible for different parts along the vertical axis of the input (note that this concept can also be applied to the horizontal direction).
The output of the \ac{slc} is then given by
\begin{equation}
\label{eq:semi-conv}
    y_{h, w, c_y} =
    \sum_{c_x=0}^{C_x-1} \sum_{i=0}^{I-1} \sum_{j=0}^{J-1}
    k_{i, j, c_x, c_y, \alpha_h} \cdot
    x_{h-i_I, w-j-J, c_x}
\end{equation}
where $\alpha_h = \left\lfloor \alpha \cdot h / H_x\right\rfloor$ selects the respective filter-component depending on the vertical position $h$.

For $\alpha = H_x$, there is no weight sharing along the vertical axis, a new filter is used for every single data row.
For $\alpha = 1$, we obtain a regular convolution as defined in~\eqnref{eq:conv}.
For values in between, the degree of weight sharing can be adapted to the desired application.
All equations omit padding, bias, and activation function for better readability.


\section{Experiments}
\label{sec:experiments}

In this section, we investigate the introduction of \ac{slc} layers in various experiments.
We use the implementation of Milioto~\etal~\cite{Milioto2019IROS}\footnote{Code: https://github.com/PRBonn/lidar-bonnetal} for our experiments.
All results are reported on the validation split of the SemanticKITTI dataset~\cite{Behley2019ICCV}.
The input to the network is a two-channel image with the projected depth measurements and the respective reflecivity values.

\tabref{tab:results}~shows the overall results for two different base networks.
First, it shows that \acp{slc} do not outperform normal convolutions.
Second, the performance decreases with increasing $\alpha$.
Third, the above two points apply to both base networks, even though DarkNet21 has approximately 24.7M trainable parameters, whereas SqueezeSegV2 only has approximately 928.5k parameters.

\begin{table}
    \centering
    \caption{
        \textbf{Overall Results}:
        Shown is the semantic segmentation performance over 19 object classes for two different base networks on the validation split of the SemanticKITTI dataset.
        Each base network is augmented with our \ac{slc} layer and tested for different values of $\alpha$, \ie the number of vertical filters within the convolution.
        All convolution layers, except input and output layer, are replaced within the network.
    }
    \label{tab:results}
    \begin{tabularx}{\linewidth}{X X ccccccc}
        \toprule
        \multirow{2}{*}{Network} & \multirow{2}{*}{Metric} & \multicolumn{7}{c}{\# vertical filters $\alpha$} \\
        \cline{3-9}
        & & 1 & 2 & 4 & 8 & 16 & 32 & 64 \\
        \midrule
        \multirow{2}{*}{DarkNet21~\cite{Milioto2019IROS}} & Accuracy & \textbf{0.83} & 0.82 & 0.81 & 0.79 & 0.75 & 0.73 & - \\
        & mIoU & \textbf{0.36} & 0.34 & 0.32 & 0.30 & 0.27 & 0.26 & - \\
        \midrule
        \multirow{2}{*}{SqueezeSegV2~\cite{Wu2019ICRA}} & Accuracy & \textbf{0.84} & 0.82 & 0.81 & 0.80 & 0.76 & 0.75 & 0.71 \\
        & mIoU & \textbf{0.36} & 0.35 & 0.31 & 0.30 & 0.27 & 0.26 & 0.25 \\
        \bottomrule
    \end{tabularx}
\end{table}

When $\alpha$ increases with a factor of $2$, then also the number of trainable parameters increases by a factor of $2$ (approximately).\footnote{DarkNet21 with $\alpha=64$ is too large for a single GPU, therefore no results are reported}
Therefore, we conduct a second row of experiments, where we reduce the number of trainable parameters in the base network by decreasing the number of filters in each layer.
Naturally, we expect a network with lower capacity to perform worse.
\tabref{tab:scaled}~shows the semantic segmentation accuracy and mIoU performance for the DarkNet21 model.
The entries with the gray background mark those networks that have approximately the same number of trainable variables as the base network, since the modification in $\alpha$ and output filter size cancel each other out.
Here again, an increase in $\alpha$ leads to decreased performance.
One has to note, that even if the number of parameters is the same for the gray cells, the increase in $\alpha$ only leads to more capacity over the spatial dimension of the feature maps, whereas larger output filters in general lead to more capacity over the depth of the network for the entire spatial extent.

\begin{table}
    \centering
    \caption{
        \textbf{Scaled Network Performance}:
        This table shows the semantic segmentation performance of the DarkNet21~\cite{Milioto2019IROS} model (Accuracy / mIoU).
        Over the columns, we increase the number of vertical filters within each \ac{slc} layer.
        Over the rows, we decrease the number of overall output filters for each layer, \ie 2 means multiplying the number of filters by a factor of $\frac{1}{2}$ which results in $\frac{1}{4}$ of the original trainable variables.
        The gray cells mark those that contain configurations where the increase in $\alpha$ is neutralized with the decrease of filter sizes and thus results in approximately the same number of trainable parameters.
    }
    \label{tab:scaled}
    \begin{tabularx}{\linewidth}{X c@{\hskip6pt}c@{\hskip6pt}c@{\hskip6pt}c@{\hskip6pt}c@{\hskip6pt}c@{\hskip6pt}c}
        \toprule
        & 1 & 2 & 4 & 8 & 16 & 32 & 64 \\
        \midrule
        1 & \colorbox{black!25}{0.83 / 0.36} & 0.82 / 0.34 & 0.81 / 0.32 & 0.79 / 0.30 & 0.75 / 0.27 & 0.73 / 0.26 & - \\
        2 & 0.83 / 0.35 & 0.81 / 0.32 & \colorbox{black!25}{0.83 / 0.33} & - / - & - / - & - / - & 0.70 / 0.24 \\
        4 & 0.82 / 0.34 & - / - & 0.77 / 0.30 & 0.76 / 0.28 & \colorbox{black!25}{0.74 / 0.25} & - / - & 0.72 / 0.24 \\
        8 & 0.77 / 0.31 & - / - & - / - & 0.74 / 0.27 & 0.72 / 0.25 & 0.71 / 0.24 & \colorbox{black!25}{0.71 / 0.23} \\
        \bottomrule
    \end{tabularx}
\end{table}

\section{Discussion}
\label{sec:discussion}

The experiments show that \acp{slc} are not able to outperform normal convolutions and performance usually decreases with increasing~$\alpha$.
This effect is stronger for networks with a large number of parameters.
Therefore, we assume that normal convolution layers of adequate capacity can already handle the different statistical properties across the vertical spatial dimension.
This claim is supported by the findings of Kayhan~\etal~\cite{Kayhan2020CVPR} who show that convolutional layers exploit absolute spatial location.
Therefore CNNs are in fact not translation invariant which means a network with sufficient capacity is able to learn even differing statistics over the vertical dimension of such a LiDAR scan.

\section{Limitations and Future Work}
\label{sec:limitations}

In contrast to other works, we present an independent ablation on our proposed \ac{slc} layer.
However, we still see the need for more extensive evaluations to obtain comprehensive evidence of the effects of \acp{slc}.
First, is must be clarified whether the results are data dependent.
This can be achieved by using other datasets, such as nuScenes~\cite{Caesar2020CVPR} and Waymo Open~\cite{Sun2020ArXiv}, or dense depth representations instead of range images.
Second, the reported results could be task-dependent and  therefore \ac{slc} might yield different results for tasks, such as object detection or motion estimation.

Future work may include a soft version of the proposed \ac{slc}, where the $\alpha$ strips of data are not processed completely independently.
The output is then computed via a linear combination of multiple kernel operations, weighted with the vertical position of the respective filter.


\section{Conclusion}
\label{sec:conclusion}

This paper presented \acfp{slc}, a network layer that only has limited weight sharing along the vertical spatial dimension of the input.
Our experiments showed that using \acp{slc} instead of normal convolutions decreases segmentation performance, especially when decreasing the amount of weight sharing.
Since normal convolution layers can already exploit spatial location information within the network, it is not necessary to explicitly address the large difference in appearance along the vertical axis of a LiDAR scan in a special layer.

{\small
\bibliographystyle{bibliography/ieee_fullname}
\bibliography{bibliography/journals_long,bibliography/refs}
}





\end{document}